\documentclass[conference,a4paper]{APSIPA2026}
\usepackage{amsmath}
\usepackage{graphicx}
\usepackage{multirow}
\usepackage{threeparttable}
\usepackage[backend=biber,style=ieee,]{biblatex}
\usepackage{geometry}
\usepackage{fancyhdr}

\fancypagestyle{firststyle}{
  \fancyhf{}
  \fancyhead[C]{2026 Asia Pacific Signal and Information Processing Association Annual Summit and Conference (APSIPA ASC)}
}

\usepackage{booktabs}
\usepackage{amssymb}
\usepackage{balance}
\begin{document}

\title{Language-Oriented Semantic Modeling for Target Language Extraction}

\author{
\authorblockN{
Mehmet Sinan Yıldırım\authorrefmark{1},
Ke Zhang\authorrefmark{2},
Shuai Wang\authorrefmark{3},
Haizhou Li\authorrefmark{1}\authorrefmark{2}
}

\authorblockA{
\authorrefmark{1}
National University of Singapore, Singapore \\
E-mail: sinan@u.nus.edu}

\authorblockA{
\authorrefmark{2}
The Chinese University of Hong Kong, Shenzhen, China \\
}

\authorblockA{
\authorrefmark{3}
Nanjing University, Suzhou, China \\
}

}
\newcommand{\blue}[1]{\textcolor{blue}{#1}}
\newcommand{\red}[1]{\textcolor{red}{#1}}

\maketitle
\thispagestyle{firststyle}
\pagestyle{empty}

\begin{abstract}
    Target Language Extraction (TLE) aims to isolate speech in a target language from multilingual mixtures. Existing approaches primarily rely on acoustic modeling with language identity labels, which provide only coarse categorical guidance. In this work, we propose a language-oriented semantic modeling framework that augments acoustic extraction with language-specific semantic representations. A pre-trained speech semantic model is adapted via language-aware prompt conditioning and parameter-efficient fine-tuning to generate target-language-focused semantic representations from the mixture. {These capture high-level semantic structure of the acoustics and they} are integrated into the extraction model to provide structured semantic guidance beyond identity labels. Experiments on the CommonVoiceMix dataset show that incorporating semantic representations consistently improves over acoustic-only baselines, achieving up to 1.16 dB SI-SDR gain. 
\end{abstract}

\begin{IEEEkeywords}
  Language-Conditioned Speech Separation, Semantic Modeling, Prompt Adaptation, Multilingual Speech Separation 
\end{IEEEkeywords} 

\section{Introduction}
An auditory landscape often contains signals from multiple sources. Humans have the remarkable ability to selectively attend to a single source while suppressing others defined as the Cocktail Party Problem~\cite{cherry1953cocktail}. Replicating this in machines is crucial as a pre-processing step for downstream single-source applications, yet isolating specific signals from multi-source mixtures still poses a great challenge. 

The extraction process requires an auxiliary cue to define the target source. Target Speaker Extraction (TSE) typically uses a voiceprint as the cue that is obtained from an enrollment utterance by the target speaker~\cite{ecapatdnn, voicefilter, tseoverview, speakerbeam, spexplus}. Other effective cues include text prompts about the speaker~\cite{textguidedtse, textguidedtse2} or the content~\cite{jiang2024ptse}, visual cues such as facial images~\cite{facetse}, facial videos~\cite{lookingtolisten} or lip movements~\cite{avsepformer} and even contextual history~\cite{contextualspeechextraction} or EEG signals~\cite{eegtse, neurospex, neuroheed}.

A similar scenario arises in multilingual environments where speech is extracted based on the spoken language, i.e. the language serves as the cue. This presents itself commonly in international environments such as airports and academic conferences. Unlike the practically infinite variability of individual speakers, languages are governed by finite and unique phonotactic, syntactic and linguistic characteristics that can be explicitly modeled. In addition, downstream models for a specific language are specialized on this unique nature. As such, it is crucial to separate audio mixtures into individual languages. This is called Target Language Extraction (TLE) first introduced in~\cite{borsdorf2021target1}.

TLE has been studied in a limited scope. Previous works~\cite{borsdorf2022target2, borsdorf2022target3} adapted TSE architectures for TLE such as ConvTasNet~\cite{convtasnet}, Spex+~\cite{spexplus}, Sepformer~\cite{subakan2021sepformer} which primarily operate on acoustic features to perform the extraction in a latent space. These acoustic-centric approaches fail to exploit the higher-level semantic structure of the underlying language.

\cite{psychology1} shows that the encoding of natural language in the human cognitive system is affected significantly by the semantic expectation of the listener. Moreover, semantic context helps listeners sharpen their comprehension in noisy environments~\cite{biology1} enhancing their understanding. 
{Motivated by this, we propose to incorporate high-level structural representations into the TLE pipeline to guide the extraction.}

Recent advances in large language models provide access to more fine-grained semantic representations. These architectures model the deep semantic relationships in a language which can be leveraged for different speech tasks such as speech enhancement \cite{li2025saslm}, speech emotion recognition \cite{mai25b_interspeech_aasllm} or generative TSE \cite{tselm}. {More directly, language models have been applied to cocktail-party tasks, e.g. LauraTSE \cite{tang2025lauratse} for target speaker extraction and MT-LLM \cite{meng2025mtllm} for multi-talker transcription. Unlike these, which use the language model as the extraction engine itself, we employ it as semantic guidance for a separate acoustic backbone.} However, the integration of discrete semantic priors into the extraction pipelines remains unexplored.

We propose a language-oriented semantic modeling approach where a large speech model is trained to predict clean semantic tokens from multilingual mixtures via language-specific prompt-conditioning and injecting them into the language extraction pipeline to guide the extraction. Context tokens for different languages condition the large speech model to generate semantic tokens for the specific target language. 
{Unlike \cite{yildirim2025apsipa}, which guides extraction using coarse language-identity cues with a dedicated model per language, we integrate generated semantic tokens into a single shared TLE pipeline that operates on the acoustic space, providing structured guidance beyond categorical language choice.}
Experiments on the CommonVoiceMix~\cite{yildirim2025apsipa} show that this approach improves the extraction quality over the baseline by up to 1.16 dB SI-SDR.

\begin{figure*}
    \centering
    \includegraphics[width=\linewidth]{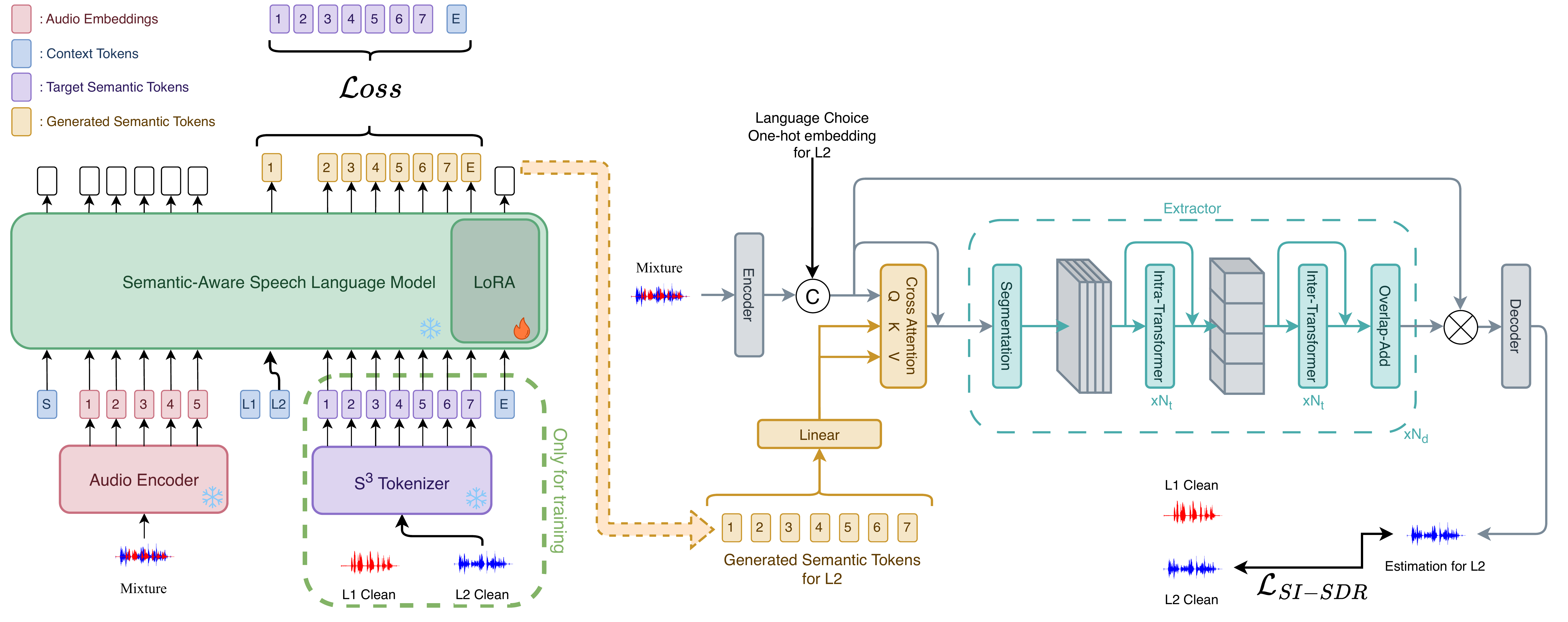}
    \caption{The general overview of the proposed approach. The left half is the language-oriented semantic modeling via generic speech semantic model converted into a language-focused semantic extractor via prompt conditioning and parameter-efficient adaptation. The generated semantic tokens are incorporated into the baseline extraction architecture via cross-attention on the right half. L2 means language 2 present in the mixture.}
    \label{fig:overview}
\end{figure*}

To sum up, our contributions are:
\begin{itemize}
    \item We propose a language-oriented semantic adaptation framework for Target Language Extraction, where a pretrained speech semantic model is conditioned via prompt design and LoRA fine-tuning to produce target-language-focused semantic representations from multilingual mixtures.
    
    \item We incorporate the predicted semantic tokens into the separation model to guide extraction with language-structured semantic information beyond coarse language identity labels.

    \item Through systematic ablation studies, we demonstrate that language identity labels provide categorical constraints while semantic representations offer structural guidance, and their combination yields the best performance.

    \item We validate the proposed language-oriented semantic modeling on CommonVoiceMix, achieving up to 1.16 dB SI-SDR improvement over a strong TLE baseline.
\end{itemize}

\section{Methodology}
\subsection{Target Language Extraction}\label{sec:baseline}
Given a multilingual mixture $m(t)=\sum_{i=1}^Nx_i(t)$, where each $x_i(t)$ corresponds to a speech signal in one language, the goal of Target Language Extraction (TLE) is to estimate the speech associated with a specified target language $k$. This is achieved by learning an extraction function $f_\theta(\cdot)$ conditioned on the mixture and the target language identity:
\begin{align}
\widehat{x}_k(t) = f_\theta\Big(m(t), k\Big)
\end{align}

The backbone for the TLE baseline is based on the Sepformer~\cite{subakan2021sepformer} architecture. As displayed in the right half of Figure \ref{fig:overview}, the encoder transforms the mixture waveform into the latent embeddings. A one-hot vector representing the target language identity is concatenated to the mixture embeddings along the feature dimension, providing the necessary conditioning for the extraction process. Subsequently, these embeddings are processed by a dual-path transformer network which alternately models the local dependencies via the intra-transformer and the global dependencies via the inter-transformer, producing a final extraction mask. The masked embeddings are decoded to reconstruct the estimated monolingual waveform. The loss function is computed between the estimated and the ground-truth monolingual target signals in the time domain.

\subsection{Language-Oriented Semantic Modeling}
The proposed approach adopts a semantic-aware speech language model~\cite{li2025saslm} as the backbone for language-oriented semantic modeling. During training, an audio encoder processes the mixture signal to obtain audio embeddings denoted in red in Figure~\ref{fig:overview}. Similarly, a semantic-aware tokenizer processes the clean target speech waveform to acquire {discrete tokens that encode high-level, semantically-rich acoustic structure learned through large-scale pretraining, rather than low-level spectral features. We refer to these as semantic tokens throughout} and they are denoted in purple in Figure~\ref{fig:overview}.

Context tokens displayed in blue are used to combine these two sequences into a single input sequence for the language model. Start and end tokens are used to signify the start and the end. In addition, a language identity token is inserted between the audio embeddings and the semantic tokens. This represents the transition from the multilingual mixture embeddings to monolingual clean tokens. Moreover, this language token is chosen from a bank of possible language identity tokens containing one per supported language. This conditions the model to generate the tokens for the chosen language. The resulting input sequence is formatted as "$<$start$>$, mixture embeddings, $<$language$>$, clean semantic tokens, $<$end$>$". The main diagram in Figure~\ref{fig:overview} displays the case for L2 without loss of generality. 

The language model takes the mixture embeddings and the language token as a prefix and autoregressively predicts the probability distribution for the next token given all previous tokens. {The training objective is the cross-entropy loss over} the predicted semantic token positions while outputs corresponding to mixture embeddings and control tokens are ignored. The predicted semantic tokens are denoted in orange, whereas the irrelevant outputs are uncolored. 

During inference, the clean speech waveform is not accessible. Semantic tokens are generated auto-regressively starting with the prompt "$<$start token$>$, mixture embeddings, $<$language token$>$" until the model predicts the $<$end$>$ token.

\subsection{Injection of Semantic Information}
The generated semantic embeddings provide language-structured semantic guidance and they are integrated into the extraction baseline. A linear projection layer maps the semantic tokens to the latent dimensionality of the extraction encoder, aligning them with the acoustic embedding space. Subsequently, they are injected before the extractor via a cross attention block after the one-hot embedding concatenation. The acoustic mixture embeddings serve as queries, while the semantic tokens act as keys and values. A residual connection around the attention block enables the fusion of semantic information with the acoustic representations, resulting in semantically informed acoustic features. Together with the one-hot language embedding, the extractor is conditioned to leverage semantic information during latent-space extraction. The remaining operations follow the baseline described in Section \ref{sec:baseline}.

\section{Experiments}
\subsection{Experimental Setup}
The baseline architecture is based on the Sepformer~\cite{subakan2021sepformer} implemented in Speechbrain\footnote{https://github.com/speechbrain/speechbrain} with the default parameters and a single mask. The loss function is the standard Scale-invariant Signal-to-Distortion Ratio (SI-SDR)~\cite{le2019sdr}. The model is trained on the CommonVoiceMix dataset introduced in~\cite{yildirim2025apsipa} consisting of mixtures obtained from English and German utterances from the CommonVoice dataset~\cite{ardilaCommonVoiceMassivelyMultilingual2020}. We use an AdamW optimizer with an initial learning rate of $3\cdot{10}^{-4}$. The first 10 epochs serve as a warmup phase where the learning rate is increased linearly until the initial value at the 10th epoch. Afterwards, the learning rate is halved if the validation SI-SDR does not improve for three epochs. Training is terminated if the validation performance does not improve for 20 consecutive epochs. 

The speech language model is a Llama-based architecture. We use the pretrained model from \cite{li2025saslm} trained on the Emilia dataset~\cite{emiliadataset}. Similarly, the audio encoder is Whisper~\cite{radford2023whisper} and the tokenizer is the S3-Tokenizer~\cite{an2024s3tokenizer}. This allows the model to leverage semantic representations learned from large-scale pretraining. This is a large model and it is costly to train in its entirety. We apply LoRA~\cite{hu2022lora} for parameter-efficient fine-tuning~\cite{peft} for the attention blocks, specifically the Q and V projection matrices. We initialize the weights from the pretrained model\footnote{https://huggingface.co/ASLP-lab/SenSE} and train with Adam optimizer with an initial learning rate of $3\cdot{10}^{-4}$. The learning rate is decreased by 1\% after every epoch. The language-specific prompt tokens are initialized with the $<$Turn of Speech$>$ token and then learned independently from each other for each language. The cross-attention module used for semantic injection is a multi-head attention block with 8 heads. {The semantic model is trained separately first, followed by the training of the proposed extraction module.}

{The full semantic model comprises approximately 1.1 B parameters, dominated by the frozen audio encoder and the Llama backbone. Only 14.0 M parameters ($\approx$1.3\% of the total) are trainable. This keeps the adaptation cost modest despite the large backbone. The inference is done in a greedy fashion with a hard limit on maximum number of tokens.}

\subsection{Results}
The main results compare the baseline model without the semantic tokens to the proposed approach incorporating semantic guidance. The results are presented in Table~\ref{tab:main}. In addition to the SI-SDR values, we also present the STOI~\cite{taal2010stoi} and PESQ~\cite{rix2001pesq} performance. SI-SDR measures the signal reconstruction fidelity whereas STOI and PESQ indicate the objective intelligibility and perceptual quality respectively. 

The approach guided by the semantic tokens outperforms the baseline across all metrics. There is a consistent improvement of 1.27 dB and 1.16 dB on the validation and test sets, respectively, {with absolute values consistent with the baseline reported on the same data in \cite{yildirim2025apsipa}}. This indicates that the semantic tokens provide informative contextual guidance that enhances the extraction of the target language.


\begin{table}
  \caption{Extraction performance with/without semantic tokens. $\uparrow$ means higher is better.}
  \label{tab:main}
  \centering
  \begin{tabular}{ccccc}
    \toprule
    \textbf{Set} & \textbf{Method} & \textbf{SI-SDR (dB)}$\uparrow$ & \textbf{STOI} $\uparrow$ & \textbf{PESQ} $\uparrow$ \\
    \midrule\midrule
    \multirow{2}{*}{Dev} & Baseline   & 11.56 & 0.86 & 1.96 \\ 
                         & Proposed       & \textbf{12.83} & \textbf{0.88} & \textbf{2.10} \\ 
    \midrule
    \multirow{2}{*}{Test} & Baseline  & 10.85 & 0.84 & 1.95 \\ 
                          & Proposed      & \textbf{12.01} & \textbf{0.86} & \textbf{2.07} \\ 
    \bottomrule
  \end{tabular}
\end{table}

\subsubsection{Results for each language}
Dataset distributions are frequently imbalanced across languages. Widely spoken languages such as English and Mandarin typically have more training data available. The pretraining was performed on the Emilia dataset~\cite{emiliadataset} that has 46800 hours of English speech and only 1600 hours of German speech. To investigate if this imbalance leads to biased representations that favor English, we present the results separated by the target language in Table~\ref{tab:reslang}. We see that the semantic tokens improve SI-SDR by 1.17 dB and 1.14 dB for English and German, respectively. The performance gains for both languages are similar. This indicates that the semantic tokens provide effective contextual guidance for both English and German. This supports the robustness of our approach suggesting that the proposed approach generalizes beyond high-resource languages and remains effective even when pretraining data is limited.

{We additionally compare against \cite{yildirim2025apsipa}, which trains a dedicated expert model per language on the same train/dev/test split. Our proposed approach outperforms these per-language experts on SI-SDR for both languages, despite using a single shared model rather than one expert per language. The gains are consistent across STOI as well, with a marginal exception on English PESQ where the two methods are comparable.}


\begin{table}
  \caption{Results for each language on test set. $\uparrow$ means higher is better.}
  \label{tab:reslang}
  \centering
  \begin{tabular}{ccccc}
    \toprule
    \textbf{Target} & \textbf{Method} & \textbf{SI-SDR (dB)}$\uparrow$ & \textbf{STOI} $\uparrow$ & \textbf{PESQ} $\uparrow$ \\
    \midrule\midrule
    \multirow{3}{*}{EN} & Baseline   & 10.72 & 0.83 & 1.90   \\
                        & \cite{yildirim2025apsipa}   & 11.18 & 0.84 & \textbf{2.05}   \\
                        & Proposed       & \textbf{11.89} & \textbf{0.85} & 2.02   \\ \midrule
    \multirow{3}{*}{DE} & Baseline   & 10.99 & 0.86 & 1.99   \\
                        & \cite{yildirim2025apsipa}   & 11.15 & 0.87 & 2.05   \\
                        & Proposed       & \textbf{12.13} & \textbf{0.87} & \textbf{2.12}   \\ 
    \bottomrule
  \end{tabular}
\end{table}

\subsubsection{Ablation of language choice with one-hot embedding}
The baseline model relies on the one-hot embedding to specify which language to extract whereas the proposed approach has access to both the one-hot embedding and the semantic tokens. To investigate if the semantic tokens from the target language are enough for the model to determine the target language, we conducted the experiment without the one-hot embedding and with only the semantic tokens. The results are presented in Table~\ref{tab:onehot}. 

Providing semantic tokens alone yields results inferior to the baseline, suggesting that these tokens lack the discriminative power to disambiguate the target source without categorical identity labels. A plausible explanation is that, without the explicit language identity input, the model may fail to reliably determine the target language and occasionally extract speech from the interfering source, leading to degraded performance. This indicates that explicit language identity conditioning remains essential, while semantic tokens provide complementary fine-grained contextual guidance. This aligns with the expectation that the semantic tokens provide semantic information only.


\begin{table}
  \caption{Ablation of the one-hot language embedding. $\uparrow$ means higher is better.}
  \label{tab:onehot}
  \centering
  \begin{tabular}{ccccc}
    \toprule
    \textbf{One-hot} & \textbf{Tokens} & \textbf{SI-SDR (dB)}$\uparrow$ & \textbf{STOI} $\uparrow$ & \textbf{PESQ} $\uparrow$ \\
    \midrule\midrule
    $\checkmark$ & $\times$ & 10.85 & 0.84 & 1.95 \\ 
    $\times$         & $\checkmark$ & 9.14 & 0.82 & 1.89  \\ 
    $\checkmark$         & $\checkmark$   & \textbf{12.01} & \textbf{0.86} & \textbf{2.07} \\ 
    \bottomrule
  \end{tabular}
\end{table}

\subsubsection{Contradictory Language Input}
Previous results suggest that the semantic tokens are not sufficient and the target language choice remains crucial. To verify that the main choice of the target language is governed by the one-hot embedding, we evaluate configurations where either the one-hot embedding or the semantic tokens are replaced with those corresponding to the interfering language as seen in Table~\ref{tab:reverse}. Contradictory conditioning leads to severe degradation, as the model is encouraged to extract the interfering language instead of the target. Notably, reversing the one-hot embedding while keeping the semantic tokens correct results in catastrophic degradation indicating that the one-hot embedding exerts dominant control over language selection. The results confirm that the explicit one-hot embedding serves as the primary gating mechanism while the semantic tokens act in complementary capacity. However, incorrect semantic tokens still {cause a clear degradation} showing that their contribution is still essential.



\begin{table}
  \caption{Contradictory language input results. $\uparrow$ means higher is better. $\checkmark$ means correct language input whereas $\rightleftharpoons$ means mismatched language conditioning}
  \label{tab:reverse}
  \centering
  \begin{tabular}{ccccc}
    \toprule
    \textbf{One-hot} & \textbf{Tokens} & \textbf{SI-SDR (dB)}$\uparrow$ & \textbf{STOI} $\uparrow$ & \textbf{PESQ} $\uparrow$ \\
    \midrule\midrule
    $\checkmark$ & $\checkmark$ & \textbf{12.01} & \textbf{0.86} & \textbf{2.07} \\ 
    $\checkmark$  & $\rightleftharpoons$ & 2.57 & 0.71 & 1.76 \\ 
    $\rightleftharpoons$  & $\checkmark$ & $-$20.10 & 0.40 & 0.37 \\ 

    \bottomrule
  \end{tabular}
\end{table}

\subsubsection{Smaller models}
Smaller architectures often lack the modeling capacity to capture complex acoustic-only feature mappings from limited data. Therefore, smaller models may benefit from the integration of the semantic information more. We test this by halving the size of the extraction model, i.e. setting $N_t=4$ instead of $N_t=8$ in Figure \ref{fig:overview}. Table~\ref{tab:small} summarizes the performance of the smaller model. Notably, the performance gain increases to 1.6 dB for the smaller model, a {notable} jump compared to the 1.16 dB gain observed for the original variant. This suggests a promising paradigm for resource-constrained edge applications, where a compact extraction backbone is enhanced by high-fidelity semantic representations from a larger, potentially shared, language model.


\begin{table}
  \caption{Results for the model with half the size, i.e. $N_t=4$. $\uparrow$ means higher is better.}
  \label{tab:small}
  \centering
  \begin{tabular}{ccccc}
    \toprule
    \textbf{Method} & \textbf{SI-SDR (dB)}$\uparrow$ & \textbf{STOI} $\uparrow$ & \textbf{PESQ} $\uparrow$ \\
    \midrule\midrule
    Baseline    & 8.99 & 0.82 & 1.74 \\ 
    Proposed        & 10.59 & 0.84 & 1.85 \\ 
    \bottomrule
  \end{tabular}
\end{table}

\section{Conclusion}
This work presented a novel framework that bridges the gap between low-level acoustic signal processing and {high-level semantic speech representations} for Target Language Extraction. By adapting a pre-trained semantic speech model via language-aware prompt conditioning, we generated structured {semantic} priors that {substantially} enhance a time-domain extraction backbone. The resulting large speech model generates semantic tokens corresponding to a single language from a given mixture. These tokens are leveraged by a cross attention mechanism to guide the Target Language Extraction. The resulting 1.16 dB SI-SDR improvement shows that the simple extraction models predominantly rely on acoustic information and the novel approach is able to leverage the valuable semantic information for the extraction. Our ablation studies revealed a hierarchical conditioning mechanism. While categorical identity labels are essential for target disambiguation, semantic sequences provide the necessary structural guidance for high-fidelity reconstruction. In the future, we plan to extend these experiments to include other baseline architectures to show that the semantic information is beneficial irrespective of the underlying model architecture. We also aim to evaluate more typologically diverse language pairs and {to quantify the inference cost introduced by the semantic model.}









\balance
\printbibliography

\end{document}